# MOSQUITO-NET: A deep learning based CADx system for malaria diagnosis along with model interpretation using GradCam and class activation maps.

Aayush Kumar, Sanat B Singh, Suresh Chandra Satapathy, Minakhi Rout

School of Computer Engineering
Kalinga Institute of Industrial Technology Deemed to be University, Bhubaneswar, Odisha, India

{aayushk61466, sanat.b.singh99, sureshsatapathy, minakhi.rout }@gmail.com

## *Abstract*

Malaria is considered as one of the deadliest diseases in today's world which causes thousands of deaths per year. The parasites responsible for malaria are scientifically known as *Plasmodium* which infect the red blood cells in human beings. The parasites are transmitted by a female class of mosquitos known as *Anopheles.* The diagnosis of malaria requires identification and manual counting of parasitized cells by medical practitioners in microscopic blood smears. Due to unavailability of resources, its diagnostic accuracy is largely affected in large scale screening. State of the art Computer aided diagnostic techniques based on deep learning algorithms such as CNNs, with end to end feature extraction and classification have widely contributed to various image recognition tasks. In this paper, we evaluate the performance of a custom made convnet Mosquito-Net, to classify the infected and uninfected cells for malaria diagnosis which could be deployed on edge and mobile devices owing to its fewer parameters and less computation power. Therefore, it can be wildly preferred for diagnosis in remote and country side areas where there is a lack of medical facilities.

Statistical analysis demonstrate that the proposed model achieves the same accuracy as pretrained CNNs model used previously in the literature works for the malaria diagnosis, despite being 10 times lighter in terms of parameters and training time. The proposed model achieves an accuracy of 95% on validation set.

1. Introduction

Malaria is a mosquito-borne infectious disease caused by Plasmodium parasites which is spread and transmitted by female Anopheles mosquitoes. Plasmodium parasites namely P. falciparum, P. ovale, P. vivax and P. malariae are different kinds of parasites that infect humans by infecting their RBCs (red blood cells) and cause deadly life-threatening conditions, among them the effects of P. falciparum and P. vivax are the greatest threats. In 2018, 50% of estimated world's population was at risk of Malaria and in many WHO regions like South-East Asia (50%), the Western Pacific (65%) and the Eastern Mediterranean (71%) most Malaria cases were accounted due to presence of P. falciparum [1] (WHO– Malaria Key Facts). P. vivax was responsible for 75% cases in WHO region of the Americas and 47% cases in India. The WHO report [2] stated that an estimated 405,000 deaths were reported from Malaria globally compared to estimated 416,000 deaths in 2017. 67% of the estimated death counts were of children under 5 years of age. The WHO African region was home to 93% of malaria cases and 94% of the deaths in 2018.

[3] WHO recommends that all cases of suspected malaria be confirmed using parasite-based diagnostic testing - either microscopy (Examination of Blood Smears via microscope) or RDT - Rapid Diagnosis Testing (Antigen Testing – less cost effective in endemic region like sub-Saharan Africa). Microscopy has been regarded as one of the most reliable ways for detection. In the microscopy process the blood sample is taken and placed on a slide, then it is stained with an agent for highlighting malaria parasites in RBCs. The slide is then examined by a trained clinician under a microscope who manually counts the number of infected RBCs. According to malaria parasite counting protocol stated by the WHO, up to 5,000 cells may have to be counted manually by a clinician making it an extremely tedious and time-consuming process. The clinician needs to be a trained and competent microscopist to identify the malaria parasites, this becomes a prominent factor in manual diagnosis – accuracy impacted by factors like observer variability, large scale screening etc [4].

Computer-aided diagnostic (CADx) tools can help to reduce clinical burden by assisting with disease classification and interpretation using machine learning (ML, especially Deep Learning) techniques/algorithms on microscopic blood smear images. One survey article [5] gives insights on different manual and automated diagnostic techniques for malaria detection (CADx tools) and the potential of Neural Networks and Deep Learning [6] for the task.

Convolutional Neural Network [7] (CNN, also known as Space Invariant Artificial Neural Networks-SIANN) is a type of regularized form of MLP (Multi-Layer Perceptrons), a class of deep learning which has been proven very useful in image recognition & classification, video analysis, medical image analysis & other related tasks. Generally due to limited availability of data (especially in the case of medical data) pre-trained CNN models trained on large datasets like ImageNet [8] are used to capitalize on the knowledge of generic features from the images for a target application. They have proven effective but they have some downsides like if the amount of labelled training data is less the pre-trained models they can easily overfit to the training data & the size of the models is generally quite large and computationally heavy due to which their deployment on edge devices is a difficult task.

Previous works [9] compares kernel-based algorithms like support vector machine (SVM) and CNNs performance in classification of infected and normal healthy cells. Collection of images of segmented RBCs were spliced in 3 tier format – Train, Validation and Test sets (randomly). While SVM classifier performed well with around 92% accuracy, the CNNs emerged victorious with a significant difference with 95% accuracy. Liang et al. (2017) [10] evaluated and compared the performances of pre-trained and custom CNN models in classification of infected and normal healthy cell images by performing cross validation studies. Gopakumar et al. (2018) [11] created and developed a CNN (custom) to detect presence of parasites over a focal stack of slide images, it outperformed the SVM classifier which had 91.81% MCC significantly by a large amount with an MCC score of 98.77%. [12] Rajaraman et al. (2018) used pre-trained CNNs over annotated clinical image dataset to extract features from best working layers and validated their performance in classifying infected (parasitized) and healthy normal cells statistically. The ResNet-50 achieved the highest accuracy among others of 95.9%. An ideal solution [5] for examination of blood smears under microscope for malaria diagnosis in resource-poor settings would be a small portable slide analyser in which a blood slide could be inserted and then it can perform classification and analysis and provide diagnostic output. Even though we are advancing in this direction, we still are far from a field – usable device.

The most accessible device present with one and all in the current time period is smartphones – the computation power of smartphones has grown significantly over the past few years. While the results of the mentioned studies were very good and promising they were huge in size and required significant computation power for deployment on edge devices like smartphones.

We propose a very lightweight CADx system using Convolutional neural networks (CNNs) as they have the ability to automatically extract features and learn filters on their own whereas in previous machine learning solutions, features had to be manually programmed in — for example, size, colour, the morphology of the cells thus reducing the manual feature engineering and expertise required. Though they do suffer from high variance and are often prone to over fitting, to prevent this Data Augmentation has been applied consisting of random flipping of images, resizing, color jitter etc. We were able to create a model that provides promising performance and at the same time requires way less storage for deployment on edge devices like smartphones.

The organization of the rest of the paper is described as follows: section 2 describes the preliminaries required for the study to make it easy to understand. The proposed model is depicted and described in details in section 3. The implementation results, comparison and analysis of results is discussed in the section 4. Finally, the concluding remarks has been made in section 5.

## 2. Preliminary

Machine learning algorithms application in the healthcare sector has shown their potential for aiding healthcare professionals and reducing errors in diagnosis thereby relieving some of the clinical burden. Various machine learning models before when employed to the task of diagnosis of malaria have used handcrafted feature extraction methods which are optimized for a specific probability distribution dataset and suitable for handling specific variance in dimension, orientation and position of the region of interest (ROI). In the present scenario, with the availability of large amount of data and computation power of GPUs, deep learning methods performance have exceeded the performance of various ML algorithms by automatically identifying the specific attributes from structured data responsible and doing end to end feature extraction and classification. Convolutional neural networks (CNNs–a type of Neural Network which is a part of deep learning method) in specific have shown very significant and powerful results in classification of images, recognition, localization and segmentation tasks.

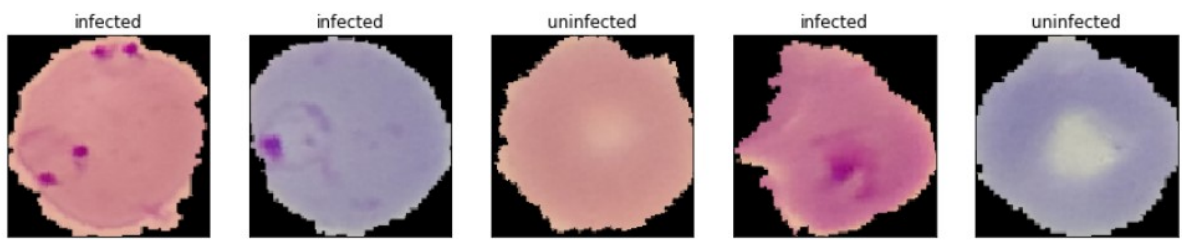

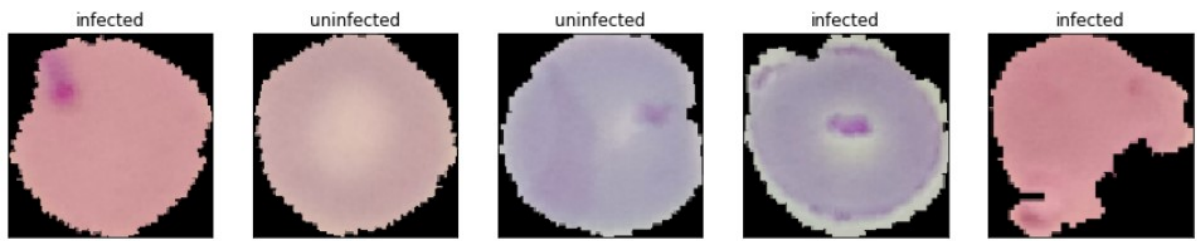

Figure 1: Example of Sample images from the infected and uninfected classes from the malaria dataset

Specifically, a class of deep learning methods, Convolutional neural networks (CNNs) have shown promising results in image classification, recognition, localization and segmentation tasks. Some sample images from the infected and uninfected classes from the malaria data set is shown in Figure 1.

The success of convolution neural networks can be attributed to the availability of huge amount of structured and annotated data. If there is shortage of data such as in medical health care fields, a different class of algorithm called transfer learning is employed which leverage the power of convolutional neural network.

## 3. Proposed Model

All the methods used so far for malaria classification have used transfer learning techniques, where models previously trained on ImageNet are used and fintuned according to the dataset requirements. The underlying principle in transfer learning is that the models transfer the weights learned during capturing the generic features from ImageNet datasets to the mentioned tasks. In 2018, Rajaraman et al. published a paper titled "Pre-trained convolutional neural networks as feature extractors toward improved parasite detection in thin blood smear images". In the mentioned paper, the authors used transfer learning techniques on five pre-trained convolutional neural networks including VGG-16 [13](Simonyan & Zisserman, 2015) , ResNet-50 [14](He et al., 2016), AlexNet [15], Xception [16] and DenseNet-121[17]. They were able to achieve impressive results but the main issue with their method is it is highly inefficient and it requires high computation power. Suppose we need to run malaria tests in a remote location where there are not enough medical facilities. There the field worker would need a device pre-loaded with the models for malaria diagnosis. The device would ideally require a light weight model which consumes less power and gives desirable accuracy in predicting malaria cases and can be easily deployed to devices incorporating Internet of Things (IOT) and edge devices.

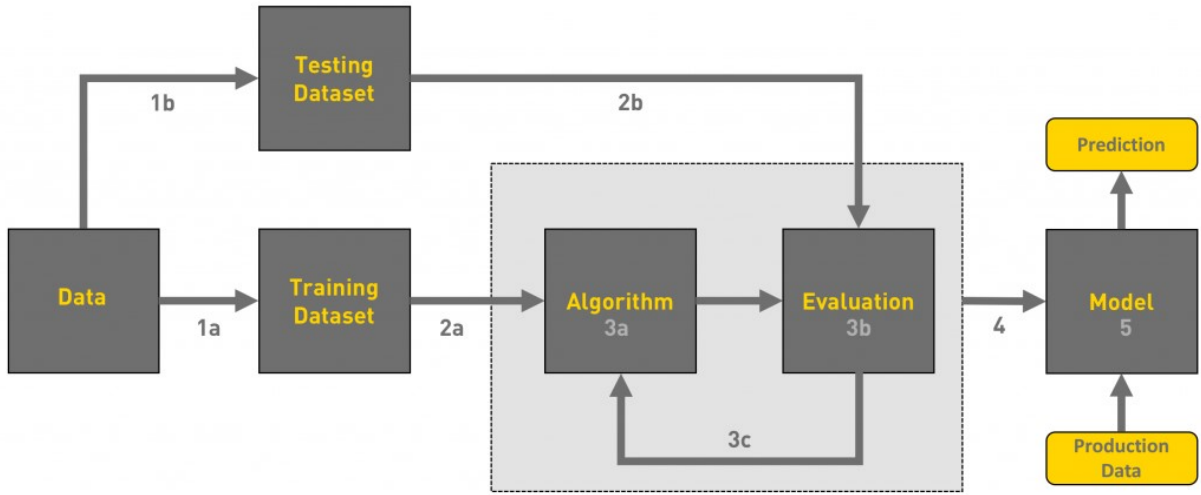

Figure-2: The proposed workflow of the Mosquito-Net is depicted in the following flow chart.

The Mosquito-Net proposed by us depicted in Figure 2 uses about 10x less time inference time on CPU than the state of the art (SOTA) ResNet50 besides being about 5x times lighter. We propose a sequential convnet (CNN) for the computer aided diagnosis system for malaria classification. The proposed Mosquito-Net shown in Figure-3 has three convolutional layers and three fully connected layers. The model is fed with an RGB image of 120 * 120 * 3 dimension. The convolutional layers use a kernel of 5 * 5 pixels and a stride of 1 pixel. Besides a padding of 2 * 2 pixels is also used. The output of a conv block is batch normalized and passed through ReLU activation function (Shang et al., 2016)[18] before being max pooled by a kernel size of 2 * 2 pixels and a stride of 2. The output of the third conv block is fed to the FC layers having 512 and 128 neurons. Dropout is used with a probability of 0.2 to avoid overfitting (Srivastava et al., 2014)[19].

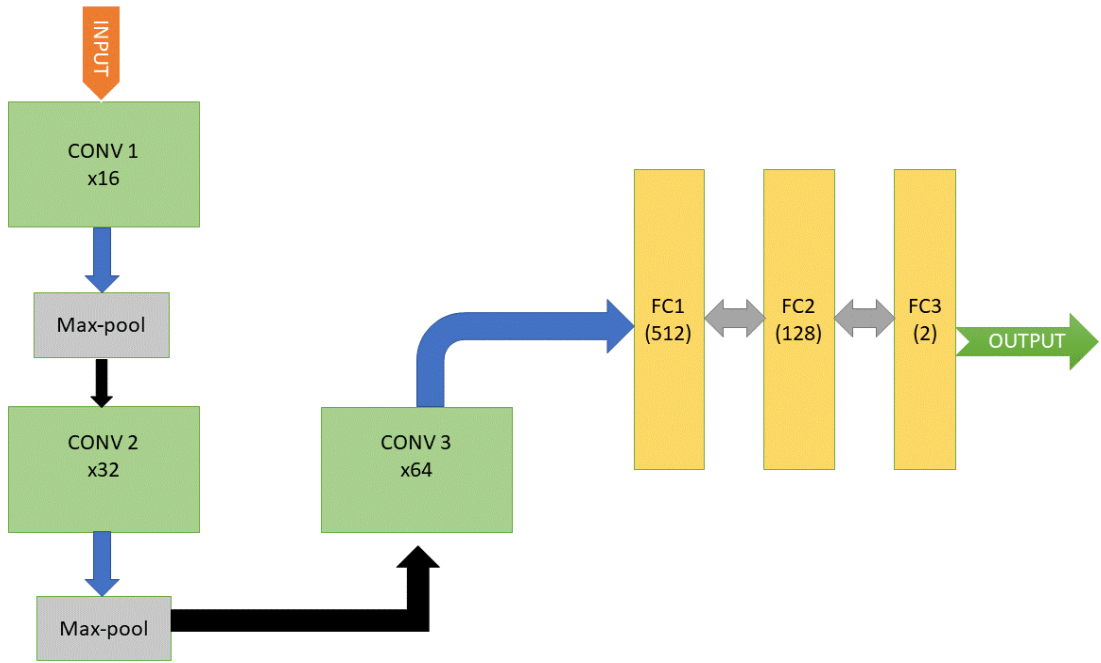

Figure-3: The proposed architecture of Mosquito-Net consisting of three convolutional blocks and three fully connected layers is depicted in the following figure

## 4. Result Analysis

Evaluation of Mosquito-Net - our custom optimized CNN model has been based on the task of classifying parasitized and uninfected cells. The hyperparameter optimization was done by manual search.

The performance of Mosquito-Net and other pretrained CNNs are evaluated in terms of the following performance metrics:

i.       Accuracy

$$Accuracy = \frac{Number\ of\ Correct\ predictions}{Total\ no\ of\ predictions\ made}$$

ii.       Precision

$$Precision = \frac{True\ Positives}{True\ Positives + False\ Positives}$$

iii.       F1-Score

$$F1 = 2 * \frac{1}{1/precision + 1/recall}$$

iv.       Sensitivity

$$Sensitivity = \frac{True\ Positive}{False\ Negative + True\ Positive}$$

v.       Specificity

$$Specificity = \frac{False\ Positive}{False\ Positive + True\ Positive}$$

vi.       Area under the receiver operating characteristic (ROC) curve

vii.       MCC

$$MCC = \frac{TP \times TN - FP \times FN}{\sqrt{(TP+FP)(TP+FN)(TN+FP)(TN+FN)}}$$

Table-1 Quantitative analysis in terms of Accuracy, AUC, F1-Score, Sensitivity, Specificity

| Models | Accuracy | AUC | Sensitivity | Specificity | F1-score | MCC |
|---|---|---|---|---|---|---|
| Alex Net | 0.927 + 0.15 | 0.978 + 0.09 | 0.939 + 0.18 | 0.931 + 0.44 | 0.939 + 0.11 | 0.869 + 0.24 |
| VGG-16 | 0.951 + 0.12 | 0.979 + 0.08 | 0.941 + 0.19 | 0.949 + 0.02 | 0.939 + 0.16 | 0.891 + 0.30 |
| ResNet-50 | 0.961 + 0.07 | 0.989 + 0.03 | 0.942 + 0.23 | 0.968 + 0.08 | 0.961 + 0.08 | 0.909 + 0.14 |
| XceptionNet | 0.889 + 0.10 | 0.956 + 0.05 | 0.941 + 0.40 | 0.841 + 0.22 | 0.889 + 0.10 | 0.781 + 0.33 |
| DenseNet-121 | 0.929 + 0.18 | 0.967 + 0.04 | 0.939 + 0.24 | 0.919 + 0.29 | 0.929 + 0.17 | 0.889 + 0.36 |
| **Mosquito Net** | **0.966 + 0.010** | **0.990 + 0.009** | **0.976 + 0.004** | **0.958 + 0.002** | **0.967 + 0.001** | **0.933 + 0.002** |

Table-2 Inference time on CPU and GPU as well as number of parameters

| Model Name | Input Size (Ch,H,W) | Params | Avg Inference time of CPU (in MS, over 100 epochs) | Avg Inference time on GPU (in MS, over 100 epochs) |
|---|---|---|---|---|
| Mosquito-Net | 3*120*120 | 7,472,002 | 0.016 | 0.002 |
| DenseNet121 | 3*224*224 | 7,978,856 | 0.327 | 0.054 |
| ResNet18 | 3*224*224 | 11,689,512 | 0.136 | 0.008 |
| ResNet34 | 3*224*224 | 21,797,672 | 0.226 | 0.015 |
| ResNet50 | 3*224*224 | 25,557,032 | 0.268 | 0.02 |
| AlexNet | 3*224*224 | 61,100,840 | 0.05 | 0.005 |
| VGG16 | 3*224*224 | 138,357,544 | 0.748 | 0.02 |
| VGG19 | 3*224*224 | 143,667,240 | 0.862 | 0.026 |
| VGG19 BN | 3*224*224 | 143,678,248 | 0.913 | 0.027 |

Evaluation of Mosquito-Net - our custom optimized CNN model has been based on the task of classifying parasitized and uninfected cells. The hyperparameter optimization was done by manual search.

Table-1 contains the mean and standard deviation values obtained for different metrics for the models over multiple cross validated folds. A reduce learning rate on plateau approach was used with respect to validation loss [20] for training of models. Mosquito-Net outperformed all other models included in the Table-1 across all metrics. The Mosquito-Net better results with 120*120 resizing of images than with 224 * 224.

The Mosquito-Net is the smallest model as seen in Table 2 with less than 8 million parameters which is closest to DenseNet-121 but took ~20x less time for inference on CPU and ~27x less inference time on GPU. AlexNet came close but was ~3x slower than Mosquito-Net on CPU inference. Table 2 states the size and execution times of model side by side. GPUs boost inference times significantly but due to their higher cost more focus is pointed on CPU inference times. However, when inference speed is a bottleneck, using GPUs provide considerable gains both from financial and time perspectives so we compared the inference times using both CPUs and GPUs which is visualized in Figure - 4. Similar conditions were maintained and checked before computing inference times for all the models. Specifications of the system are as follows: -

CPU - Intel Core i5 8300H, RAM - 8GB DDR4 RAM (2666 MHz), GPU - NVIDIA® GTX 1050Ti

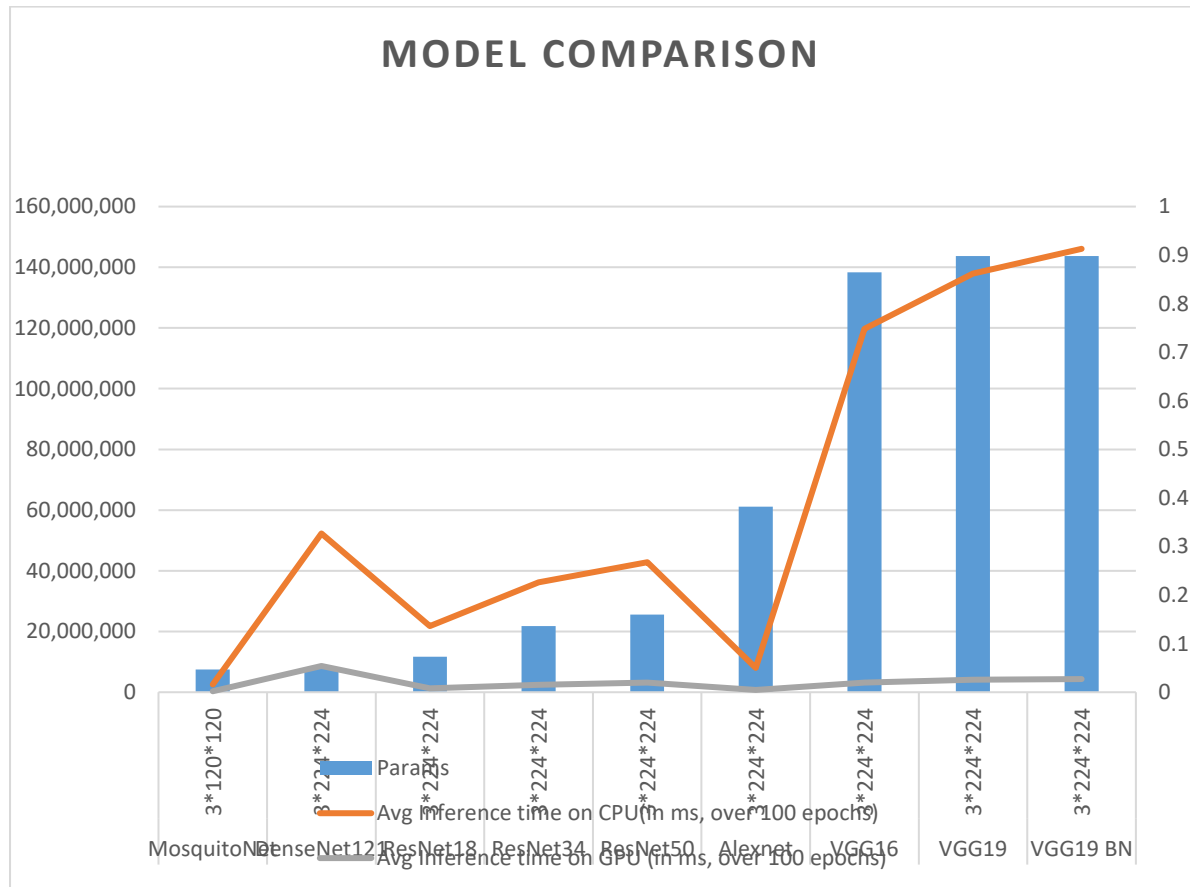

Figure 4: Depiction of average inference times of various CNN models

**Web Based Deployment**

The trained model was deployed using a web server to an Amazon EC2 instance (t2.micro instance) and was also tested out on very constrained resource instances (t3a.nano & t2.nano instances). The web server was created using Flask. Since the model being lightweight requires very less amount of memory multiple workers were able to handle simultaneous requests easily and load balancing could be done. The work flow is simple image is sent to the server via the website and the result is calculated, sent back and displayed at the client side.

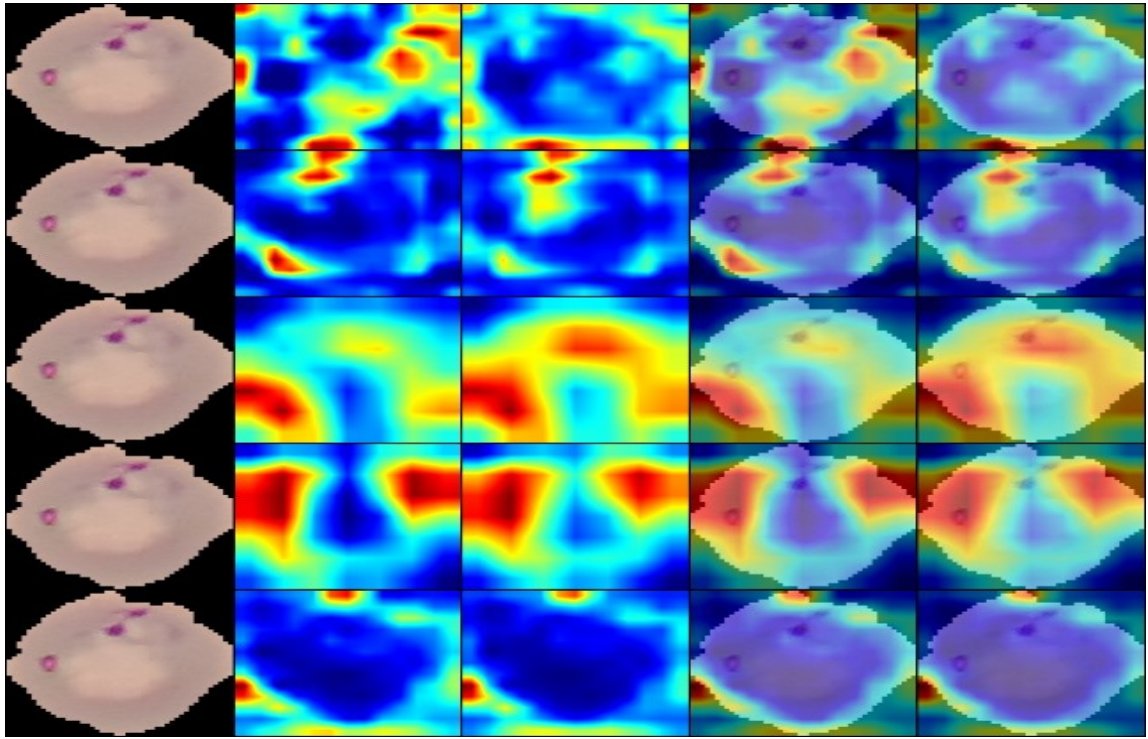

Fig: GradCam results on infected cell images.

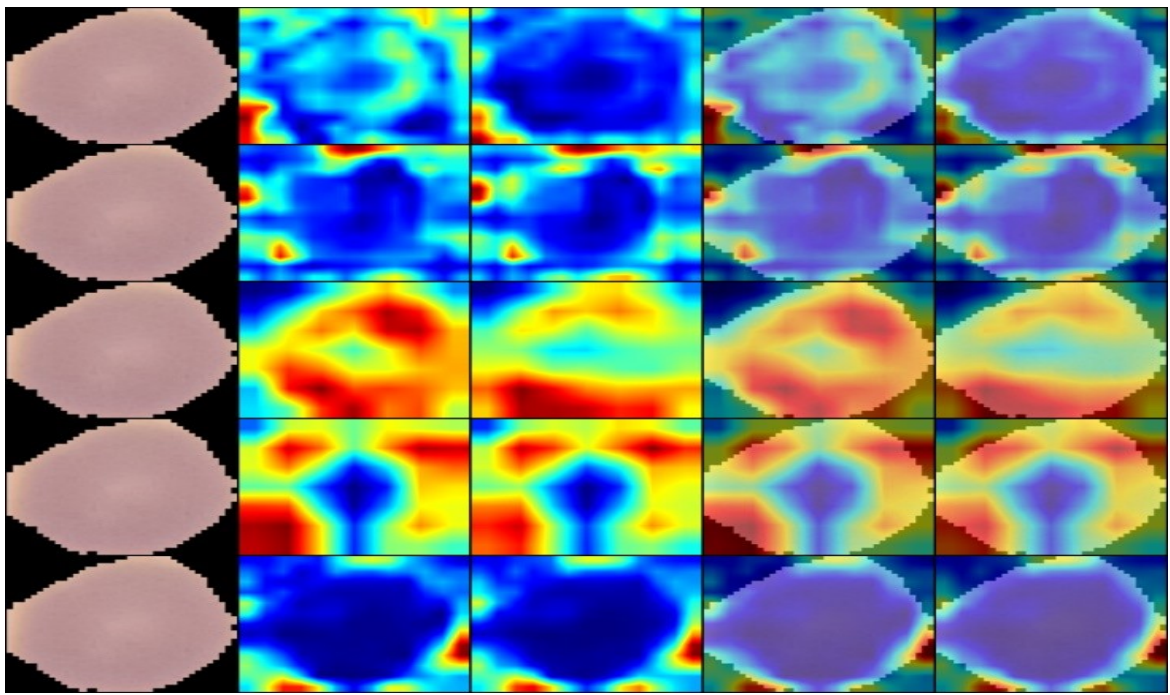

Figure-5: Example blood smear images of malaria cases for infected and uninfected samples and their associated critical factors as identified by GradCam.

## Qualitative Analysis

We further investigate and explore how Mosquito-Net makes predictions by leveraging **GRADCAM**, an explainable method that has been shown to provide good insights into how deep neural networks come to their decisions. The critical factors identified in some xample blood smear images of malaria cases are shown in Fig 5. It can be observed that the proposed Mosquito-Net identifies localized areas within the cells in the blood smeared images as being critical factors in determining whether a smear image is of a patient with a malaria infection, as shown in red in Fig 5.

## 5. Conclusion

In this study, we introduced Mosquito-Net, a deep convolutional neural network for the detection of malaria cases from blood smear images. Moreover, we investigated how Mosquito-Net makes predictions using an explainable method in an attempt to gain deeper insights into critical factors associated with Malaria cases, which can aid clinicians in improved screening as well as improve trust and transparency when leveraging Mosquito-Net for accelerated computer aided Diagnosis.

The hope is that the promising results achieved by Mosquito-Net, along with the fact that it is available in open source format will lead it to be leveraged and build upon by both researchers and citizen data scientists alike to accelerate the development of highly accuracy yet practical deep learning solutions for detecting malaria cases from blood smear images and accelerate treatment of those who need it the most.